\documentclass[conference]{IEEEtran}
\IEEEoverridecommandlockouts
\usepackage{cite}
\usepackage{amsmath,amssymb,amsfonts}
\usepackage{algorithmic}
\usepackage{graphicx}
\usepackage{textcomp}
\usepackage{xcolor}
\usepackage{svg}
\def\BibTeX{{\rm B\kern-.05em{\sc i\kern-.025em b}\kern-.08em
    T\kern-.1667em\lower.7ex\hbox{E}\kern-.125emX}}
\begin{document}

\title{End-to-End Natural Language Understanding Pipeline for Bangla Conversational Agents
% % \thanks{Identify applicable funding agency here. If none, delete this.}
}

\author{ 

\IEEEauthorblockN{Fahim Shahriar Khan\textsuperscript{1}, Mueeze Al Mushabbir\textsuperscript{1}, Mohammad Sabik Irbaz\textsuperscript{2}, MD Abdullah Al Nasim\textsuperscript{2}}
\IEEEauthorblockA{Department of Computer Science and Engineering, Islamic University of Technology\textsuperscript{1} \\
Machine Learning Team, Pioneer Alpha Ltd.\textsuperscript{2} \\
khanfahimshahriar0@gmail.com, almushabbir@iut-dhaka.edu,\\ sabikirbaz@iut-dhaka.edu, nasim.abdullah@ieee.org
}

% \IEEEauthorblockN{1\textsuperscript{st} Given Name Surname}
% \and
% \and

% \and
% \and
% \and
}

\maketitle

\begin{abstract}
Chatbots are intelligent software built to be used as a replacement for human interaction. Existing studies typically do not provide enough support for low-resource languages like Bangla. Due to the increasing popularity of social media, we can also see the rise of interactions in Bangla transliteration (mostly in English) among the native Bangla speakers.  In this paper, we propose a novel approach to build a Bangla chatbot aimed to be used as a business assistant which can communicate in low-resource languages like Bangla and Bangla Transliteration in English with high confidence consistently. Since annotated data was not available for this purpose, we had to work on the whole machine learning life cycle (data preparation, machine learning modeling, and model deployment) using Rasa Open Source Framework, fastText embeddings, Polyglot embeddings, Flask, and other systems as building blocks. While working with the skewed annotated dataset, we try out different components and pipelines to evaluate which works best and provide possible reasoning behind the observed results. Finally, we present a pipeline for intent classification and entity extraction which achieves reasonable performance (accuracy: 83.02\%, precision: 80.82\%, recall: 83.02\%, F1-score: 80\%). 
\end{abstract}


\section{Introduction}
\noindent In this era of artificial intelligence (AI), chatbots are becoming more and more popular every day for their versatility, easy accessibility, personalizing features, and, more importantly, their ability to generate automated responses \cite{nuruzzaman2018survey}. Specifically for these purposes, we now see an uprise of chatbots everywhere - from personal to organizational, to business websites or other online platforms \cite{nuruzzaman2018survey}, for which it can be trained on suitable data to make it, in a broader sense, a virtual assistant representative of the said entities.\par

\noindent In the light of this newly emerging scope, we explore the possibilities of how these conversational AI agents can be integrated properly and thus be an immensely useful tool to maintain business activities. To better understand the concurrent chatbots and find possible modifications in them and for further and more customized improvements, we choose an open-source chatbot platform Rasa as our study subject \cite{bocklisch2017Rasa}.\par

\noindent As we progress with our experiments at hand, we find that there are numerous pipeline choices available for the Rasa chatbot, and, for each of these pipelines, in turn, there are various other components in it to decide upon. One can easily find themselves in a rabbit hole in simply choosing these pipelines and their components alone. Moreover, for this vast number of choices, it is also extremely difficult to know or find out which components or pipelines are the most suitable for each use case. In this regard, while Rasa does indeed provide a default option for the users, this setting, again, is not guaranteed to be perfectly suited for most of the use cases in real-life scenarios. Additionally, among other challenges, a prominent one is having a skewed, poorly defined, or scarce dataset for training the chatbot. In our case, it turns out to be so because we are dealing with Bangla, a low-resource language. \par

\noindent In summary, our contributions include:
\begin{itemize}
  \item Creating and annotating a dataset to address Frequently Asked Questions (FAQs) by the users and their corresponding relevant replies.
  \item Performing an extensive comparative analysis among various pipelines and their consisting components.
  \item Introducing two components that work well for a low resource language such as Bangla. And while doing so, we, in turn, also suggest an overall pipeline that performs the best to solve our problem.
\end{itemize}

\begin{figure}[h]
\centering
% \includesvg[width=\columnwidth]{pipeline.eps}
% \def\svgwidth{\columnwidth}
% %\input{pipeline.pdf_tex}
\includegraphics[width=8cm]{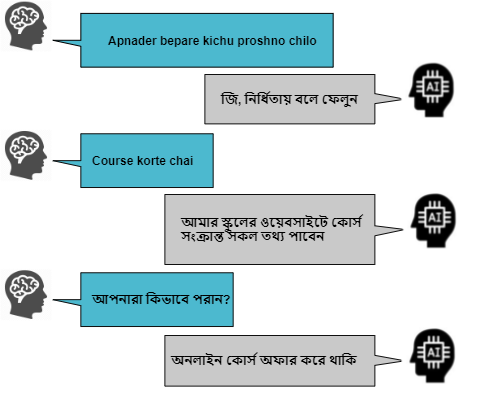}
\caption{Example of end-to-end interaction between user and bot}
\label{methodology}
\end{figure}

\section{Related Works}

\noindent In the history of conversational AI agents, ELIZA\cite{bradevsko2012survey,zemvcik2019brief}, one of the first rule-based chatbots, took it upon itself to pass the famous Turing Test and pioneer the path of guided computer responses. Even though it failed to pass the test completely, it surely did not come short in paving the way for other artificial chatbots, which ranged from responding emotionally (PARRY) \cite{bradevsko2012survey,zemvcik2019brief,abushawar2015alice} to simply having fun conversations by running pattern matching (Jabberwacky) \cite{bradevsko2012survey}. Later, this field got more matured with the inception of AI-powered chatbots, namely Dr. Sbaitso\cite{zemvcik2019brief} and A.L.I.C.E (Artificial Linguistic Internet Computer Entity)\cite{abushawar2015alice,bradevsko2012survey}- which was able to mimic humans when chatting online or answering questions. From there, it was not long before Smarterchild, Siri, Google Assistant, and other personalized assistant-like chatbots or conversational AI agents came into existence.  With conversational AI, now, anyone can build, integrate, and use message-based or speech-based assistants to automate the communication process - in personal, organizational, or business settings. And as a facilitator of this process, many conversational AI platforms have come forward and contributed largely to the advancement of the field - where one of such platforms is Rasa. \par

\noindent Rasa\cite{bocklisch2017Rasa} is an open-source python framework to build and customize conversational chatbot assistants, besides imitating humans the Rasa framework can be used to develop assistants that can perform complex tasks like booking tickets for a movie or even checking if movie tickets are available for a particular slot, etc, as the Rasa framework has functionalities to interact with databases and servers. \par

\noindent There are several chatbot frameworks other than Rasa, that are currently available and can be used to make commercial chatbots, for example, \textbf{(1)} the Microsoft Bot Framework \cite{sannikova2018chatbot}, which uses either QnA Maker service or Language Understanding Intelligent Service(LUIS) to build a sophisticated virtual assistant, \textbf{(2)} Google Dialogflow \cite{reyes2019methodology} which has the advantage of being easily integrated with all platforms virtually including home applications, speakers, smart devices, etc. On the other hand, Rasa is composed of two separate and decoupled parts \textbf{(i)} Rasa Natural Language Understanding(NLU) module to classify the intents and entities from given user input. Rasa now incorporates the DIET Architecture \cite{bunk2020diet} in the NLU pipeline which classifies both the intent and entity. \textbf{(ii)} The Rasa Core module does the job of dialogue management, which is short for deciding what actions the chatbot assistant shall take given a certain user input, intents, entities from that given input, and the current state of the conversation. Rasa uses several machine learning-based policies - like \textbf{Transformer Embedding Dialogue (TED) Policy}\cite{vlasov2019dialogue} which leverages transformers to decide which action to take next, and also rule-based policies like \textbf{Rule Policy} which handles conversation parts that follow a fixed behavior and makes predictions using rules that have been set based on NLU pipeline predictions. So in summary we can say that the Rasa framework is designed for developers with programming knowledge. \par

\noindent Furthermore, Rasa assistants can be connected with personal and business websites by building website-specific connectors for the chatbot to communicate with the client-side server. Similarly, through in-built connectors, the framework provides the functionality for integrating chatbots with Facebook pages, Slack, and other social media platforms. Virtual chatbot assistants can be developed supporting several languages by changing or even building language-specific custom components in the NLU pipeline. For example, the tokenization process for the Mandarin language will be different compared to the tokenization process of English due to language structure. As a result, the need for a language-specific tokenizer arises. In the paper by Nguyen et. al. \cite{nguyen2021enhancing}, the authors built a custom Vietnamese language tokenizer and a custom language featurizer which leveraged pre-trained fastText \cite{athiwaratkun2018probabilistic} Vietnamese word embedding and achieved a better result with their custom made components compared to the default pipeline components provided by Rasa. fastText \cite{athiwaratkun2018probabilistic} provides word embeddings for 157 languages and thus depending on the language in which the chatbot needs to be built, the corresponding featurizer, to leverage the pre-trained word vectors, must be designed and attached to the pipeline.

\section{Machine Learning Life Cycle}
\noindent Since Bangla chatbots, Bangla transliteration chatbots, or multilingual chatbots that include Bangla is not widely explored, we had to work on all the parts of the Machine Learning Life-cycle that is Dataset Preparation, Machine Learning Modeling, and Deployment for production services. Figure \ref{methodology} shows the full pipeline of the workflow.

\subsection{Dataset Preparation}
\noindent Our challenge was to deal with Frequently Asked Questions (FAQs) and some other common questions in Bangla and Bangla Transliteration in English. It was also required to provide a suitable and swift response. We collected all the FAQs from our client's interaction history with customers from different social media platforms. The questions were then annotated by 20 employees of our client with suitable answers. After the annotation process, the required files to train the model were prepared by a group of data analysts. \par

\noindent Rasa open-source architecture requires data to be split into three files. \textbf{1)nlu.yml} is required for training the NLU module and contains all the gathered FAQs categorized into 45 intents and 9 entities, with each intent containing at least 4 examples leading to around 250+ samples. \textbf{2)domain.yml} contains 150+ collected responses and actions (custom defined functions called to access and retrieve data from a database) to each corresponding FAQs in the \textit{nlu.yml} file. \textbf{3)stories.yml} contains 110+ sections, with each section containing a series of sequential \textbf{(i)} intents and entities that will be extracted from FAQs in the \textit{nlu.yml} file, \textbf{(ii)} responses and actions that can be given when a user input is categorized under a certain intent and with or without a defined entity is received, which is basically an attempt to model an actual conversation that might take place. The \textit{domain.yml} and \textit{stories.yml} files are required to train the core module which is responsible for dialogue management.

\subsection{Machine Learning Modeling} 

\noindent Although there are frameworks and platforms like Google Dialogflow, Microsoft bot framework, we use the RASA framework to build our chatbot agent. RASA is an opensource framework and much more customizable than the aforementioned two platforms and also other platforms and frameworks. For example, dialogflow can only provide one webhook for each project. This essentially means that the entire chatbot must have exactly one webhook instead of choosing multiple webhooks on an intent-by-intent basis, which is not the case for RASA, thus showcasing one of the many customizable superiorities it has over other frameworks and platforms. RASA is designed for development and consequently provides scope for a lot of customizability, whether adding specific custom components for any task-specific dataset or adding multiple webhooks for a single project and so on.

\noindent The Rasa open-source Architecture is composed of two separate units to handle a conversation with any user, \textbf{(1)} NLU: which classifies a particular sentence into a certain intent and extracts entities from it if they are present as defined in the training dataset. \textbf{(2)} Core: which decides what responses to utter or what custom actions to take if a necessary database querying is required for user input.

\subsubsection{Rasa NLU Module}
To customize the NLU pipeline which processes an input text, in the order as defined in the config file, we mainly consider three types of operations: \textbf{(1)} the choice of sparse featurizers like LexicalSyntacticFeaturizer, CountVectorFeaturizer. \textbf{(2)} choosing pre-trained dense featurizers for example a featurizer that converts tokens into dense fastText vectors. \textbf{(3)} The choice of an intent classifier and an entity extractor, for our NLU pipeline we use the DIET classifier to classify both intents and entities by using the same network.

\subsubsection{Rasa Core Module}
This module uses \textbf{(1)} a tracker which maintains the state of the conversation, \textbf{(2)} a set of rule-based and machine learning-based policies to select an appropriate response as defined in the domain file. The step, through which the core module uses to select an appropriate response is defined below:
\begin{enumerate}
\item The NLU module converts the user’s message into a structured output including the original text, intents, and entities.
\item With the output from the previous state the tracker updates its current state.
\item Policies defined uses the output from the tracker and selects an appropriate response as defined in the domain file and responds accordingly.
\end{enumerate} \par

\noindent The policies that we use for our module is discussed below.

\begin{figure}[h]
\centering
\includegraphics[width=9cm]{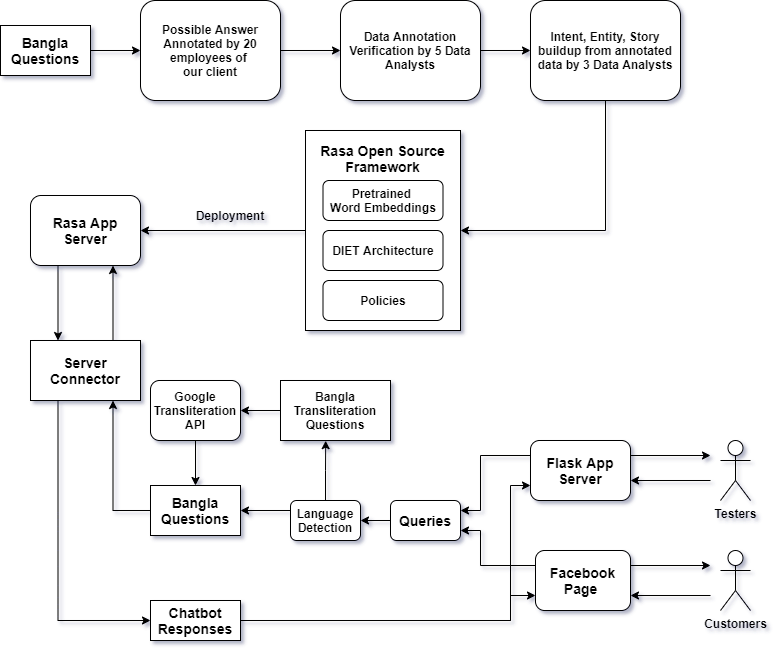}
\caption{Machine Learning Life Cycle WorkFlow}
\label{methodology}
\end{figure}

\begin{itemize}
    \item \textbf{TED Policy:} A machine learning-based architecture built on a transformer \cite{vaswani2017attention} to predict the next action to an input text from a customer.
    \item \textbf{Memoization Policy:} A rule-based policy that checks if the current conversation matches our defined stories and responds accordingly.
    \item \textbf{Rule Policy:} Used to implement bot responses for out-of-scope messages that our chatbot is not trained on so that it can fall back to a default response when the confidence values are below a defined threshold.
\end{itemize}

\subsection{Deployment} \label{development}
\noindent We deployed the trained NLU and Core module to a hosted web server. The web server can successfully take user inputs as queries and provide swift and suitable responses. We built a custom server connector so that queries from other servers can be received and responses can be sent back. To test how our hosted server is responding, we hosted another server built using Flask framework\cite{grinberg2018flask}. We also built a connection using Facebook Developer Tools \footnote{https://developers.facebook.com/tools/} between the hosted web server and a demo Facebook Page to test how our trained models work on real-life scenarios.\\

\begin{table}[h]
\caption{List of NLU Pipeline Configuration}
\label{pipelines}
\begin{center}
\begin{tabular}{|c|c|}
\hline
\textbf{\begin{tabular}[c]{@{}c@{}}NLU \\ Pipeline\end{tabular}} & \textbf{Details}                                                                                                                                                                                                                                        \\ \hline
P1                                                               & \begin{tabular}[c]{@{}c@{}}WhitespaceTokenizer + RegexFeaturizer + \\ LexicalSyntacticFeaturizer + \\ CountVectorFeaturizer + DIETClassifier\end{tabular}                                                                                               \\ \hline
P2                                                               & \begin{tabular}[c]{@{}c@{}}CustomTokenizer + RegexFeaturizer + \\ LexicalSyntacticFeaturizer + \\ CountVectorFeaturizer + DIETClassifier\end{tabular}                                                                                                   \\ \hline
P3                                                               & \begin{tabular}[c]{@{}c@{}}WhitespaceTokenizer + RegexFeaturizer + \\ LexicalSyntacticFeaturizer + \\ CountVectorFeaturizer + DIETClassifier + \\ EntitySynonymMapper +  FallbackClassifier \end{tabular}                                              \\ \hline
P4                                                               & \begin{tabular}[c]{@{}c@{}}WhitespaceTokenizer + RegexFeaturizer + \\ LexicalSyntacticFeaturizer + CountVectorFeaturizer +\\ fastTextFeaturizer +  DIETClassifier +\\ EntitySynonymMapper +  FallbackClassifier \end{tabular}                                              \\ \hline
P5                                                               & \begin{tabular}[c]{@{}c@{}}WhitespaceTokenizer + LexicalSyntacticFeaturizer + \\ CountVectorFeaturizer + fastTextFeaturizer +\\  DIETClassifier + EntitySynonymMapper +\\  FallbackClassifier\end{tabular}                                              \\ \hline
P6                                                               & \begin{tabular}[c]{@{}c@{}}CustomTokenizer + LexicalSyntacticFeaturizer +\\  CountVectorFeaturizer + \\ fastTextFeaturizer + DIETClassifier + \\ EntitySynonymMapper + FallbackClassifier\end{tabular}                                                  \\ \hline
P7                                                               & \begin{tabular}[c]{@{}c@{}}LanguageModelTokenizer (BERT) + \\ LanguageModelFeaturizer (BERT) + \\ LexicalSyntacticFeaturizer + \\ CountVectorFeaturizer  + DIETClassifier +  \\ EntitySynonymMapper + FallbackClassifier\end{tabular} \\ \hline
P8                                                               & \begin{tabular}[c]{@{}c@{}}LanguageModelTokenizer (BERT) + \\ LanguageModelFeaturizer (BERT) + \\ RegexFeaturizer + LexicalSyntacticFeaturizer + \\ CountVectorFeaturizer  + DIETClassifier +  \\ EntitySynonymMapper + FallbackClassifier\end{tabular} \\ \hline
\end{tabular}
\end{center}
\end{table}

\subsection{Interaction Between User and Bot}
\noindent After deployment, the webserver is connected with both Flask App Server and Facebook Page. Flask App Server is used by testers so that they can provide feedback and qualitative evaluation. The Facebook page deals with actual users. Whenever any user sends a message, firstly the language of the message is detected using Polyglot Word Embedding \cite{kashmir}. For now, we are only working with Bangla and Bangla Transliteration in English. To convert Bangla Transliteration in English to Bangla, we are using Google Transliteration API \footnote{https://developers.google.com/transliterate/v1/reference}. The Bangla queries pass through a custom server connector that we built for seamless interaction between Rasa App Server and end-users. The Rasa app server returns a suitable response following up the queries. 

\section{Experimental Analysis}

\subsection{Experimental Setup} 
\noindent As mentioned in Table \ref{pipelines}, we set up 8 different pipelines for our experiments, and as a part of the setting, we split our annotated data into an 80-20 ratio train-test set. For all of the experiments using different pipelines, we trained the NLU Module for 500 epochs with a learning rate of 0.05 using Adam Optimizer \cite{kingma2014adam}. In the case of the Core Module, we trained for 200 epochs with a learning rate of 0.05 using Adam optimizer \cite{kingma2014adam}.

\begin{table}[h]
\caption{Comparative Result Analysis of Intent Classification}
\label{results}
\begin{center}
\begin{tabular}{|c|c|c|c|c|}
\hline
\textbf{\begin{tabular}[c]{@{}c@{}}NLU \\ Pipeline\end{tabular}} & \textbf{Accuracy} & \textbf{Precision} & \textbf{Recall} & \textbf{F1-Score} \\ \hline
P1                                                               & 75.47             & 63.65              & 75.47           & 67.75             \\ \hline
P2                                                               & 77.36             & 68.24              & 77.36           & 70.82             \\ \hline
P3                                                               & 77.36             & 66.45              & 77.36           & 70.48             \\ \hline
P4                                                               & 73.58             & 62.74              & 73.58           & 66.49             \\ \hline
P5                                                               & 79.25             & 71.38              & 79.25           & 73.46             \\ \hline
P6                                                               & 81.13             & 73.27              & 81.13           & 75.32             \\ \hline
P7                                                               & 79.25             & 75.16              & 79.25           & 75.47             \\ \hline
P8                                                               & \textbf{83.02}    & \textbf{80.82}     & \textbf{83.02}  & \textbf{80}       \\ \hline
\end{tabular}
\end{center}
\end{table}

\noindent Among the 8 different pipelines mentioned in Table \ref{pipelines}, some include our two custom components - fastTextFeaturizer, and CustomTokenizer, which is specifically built for working with Bangla corpus. Additional arguments and parameters that we fixed for some of the components are as follows:
\begin{itemize}
  \item For LanguageModelFeaturizer, we used "bert" model configurations from HuggingFace transformers library \cite{wolf2019huggingface} as model and "Rasa/LaBSE" as model weights.
  \item For CountVectorsFeaturizer, we used "char\_wb" as an analyzer, and set 1 to be minimum ngram and 4 to be maximum ngram.
  \item For FallbackClassifier, we set 0.3 for thresholds and additionally to handle ambiguity 0.1 for ambiguity thresholds.
\end{itemize}

\noindent Lastly, for all of our experiments, we kept the policies unchanged and as following:  
\renewcommand\labelitemi{$\textendash$}
\renewcommand\labelitemii{}
\renewcommand\labelitemiii{}
\begin{itemize}
\item name: MemoizationPolicy
\item name: TEDPolicy \\
max\_history: 5 \\
epochs: 200 \\
number\_of\_transformer\_layers:
\begin{itemize}
\item text: 2
\item action\_text: 2
\item label\_action\_text: 2
\item dialogue: 2
\end{itemize}
\item name: RulePolicy
\end{itemize}

\subsection{Ablation Study}

\noindent We conducted 8 experiments to evaluate which NLU pipeline components work best for our task. The corresponding pipeline configurations are listed in Table \ref{pipelines}. Finally, we observe our obtained results and identify how and why a particular component leads to each particular result. \par

\noindent As mentioned in Table \ref{pipelines}, NLU pipeline \textbf{P1} uses \textbf{(1)} whitespace tokenizer that separates user inputs into individual tokens based on the white space that usually separates each individual word in different languages. \textbf{(2)} regex featurizer, lexical syntactic featurizer, countvector featurizer(counts character ngram tokens as features) which generates sparse features. \textbf{(3)} the DIET classifier which is trained on these features to classify intents and extract entities. And the results for the pipeline \textbf{P1} are shown in Table \ref{results}. We consider this pipeline as our baseline. \par

\noindent For pipeline \textbf{P2}, we take out the whitespace tokenizer component and replace it with our custom tokenizer specifically built for Bangla language as the tokenization process varies from language to language. As a result, we can see that test scores for accuracy, precision, recall, and F1-score increase. From this experiment, we can derive that our custom Bangla language tokenizer works better for our Bangla dataset compared to whitespace tokenizer. \par

\begin{figure}[h]
\centering
\includegraphics[width = 9cm]{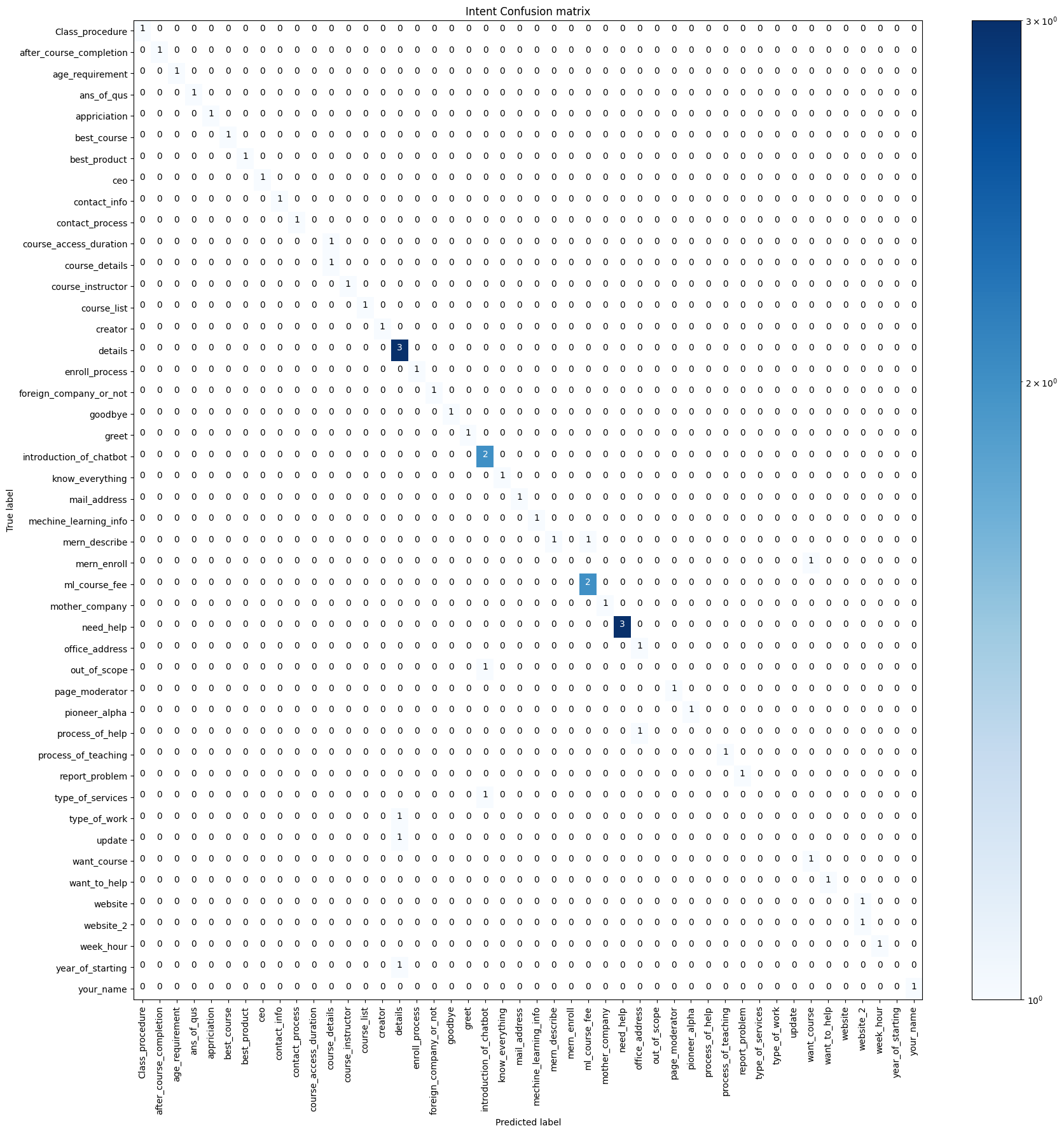}
\caption{Pipeline P6 - Confusion Matrix}
\label{p6confmat}
\end{figure}

\begin{figure}[h]
\centering
\includegraphics[width = 5.85cm]{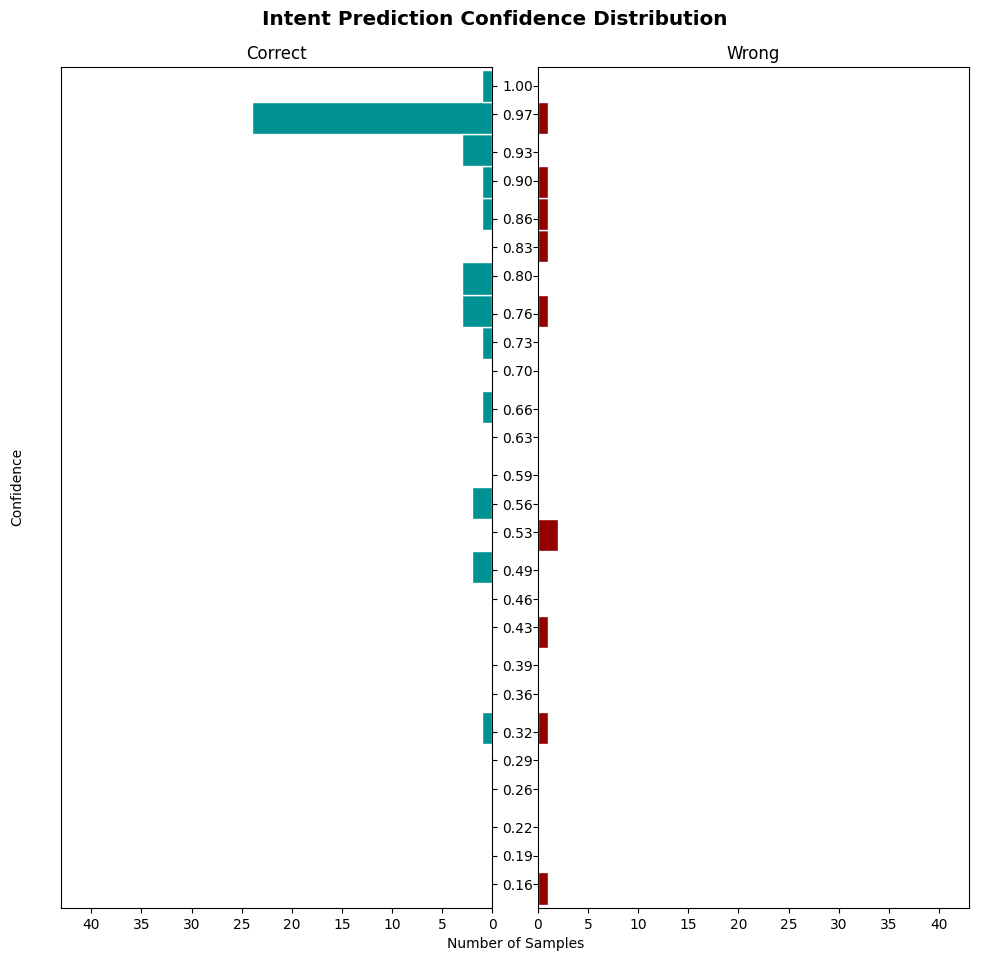}
\caption{Pipeline P6 - Intent Histogram}
\label{p6intent}
\end{figure}

\begin{figure}[h]
\centering
\includegraphics[width = 9cm]{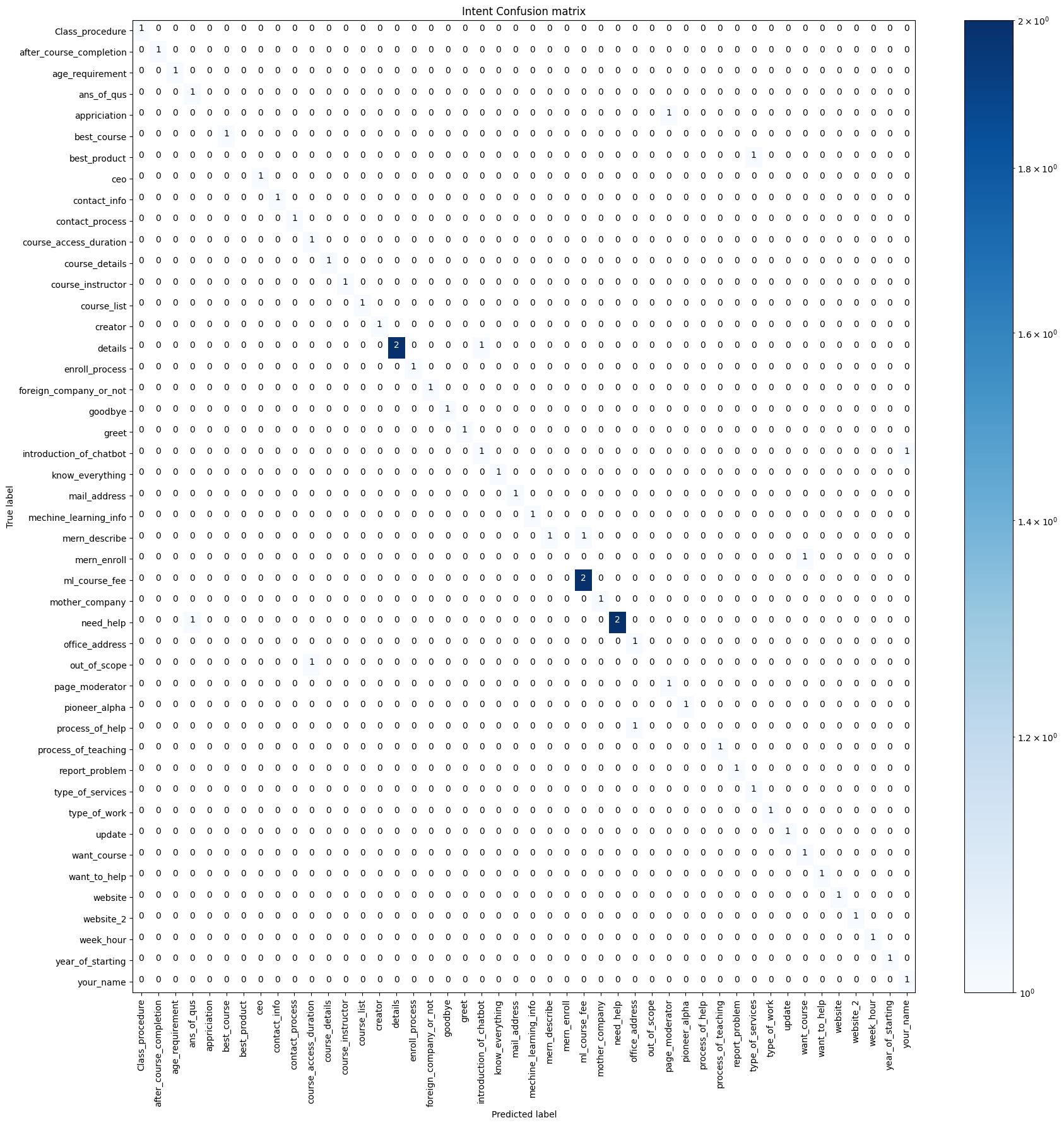}
\caption{Pipeline P8 - Confusion Matrix}
\label{p8confmat}
\end{figure}

\begin{figure}[h]
\centering
\includegraphics[width = 6cm]{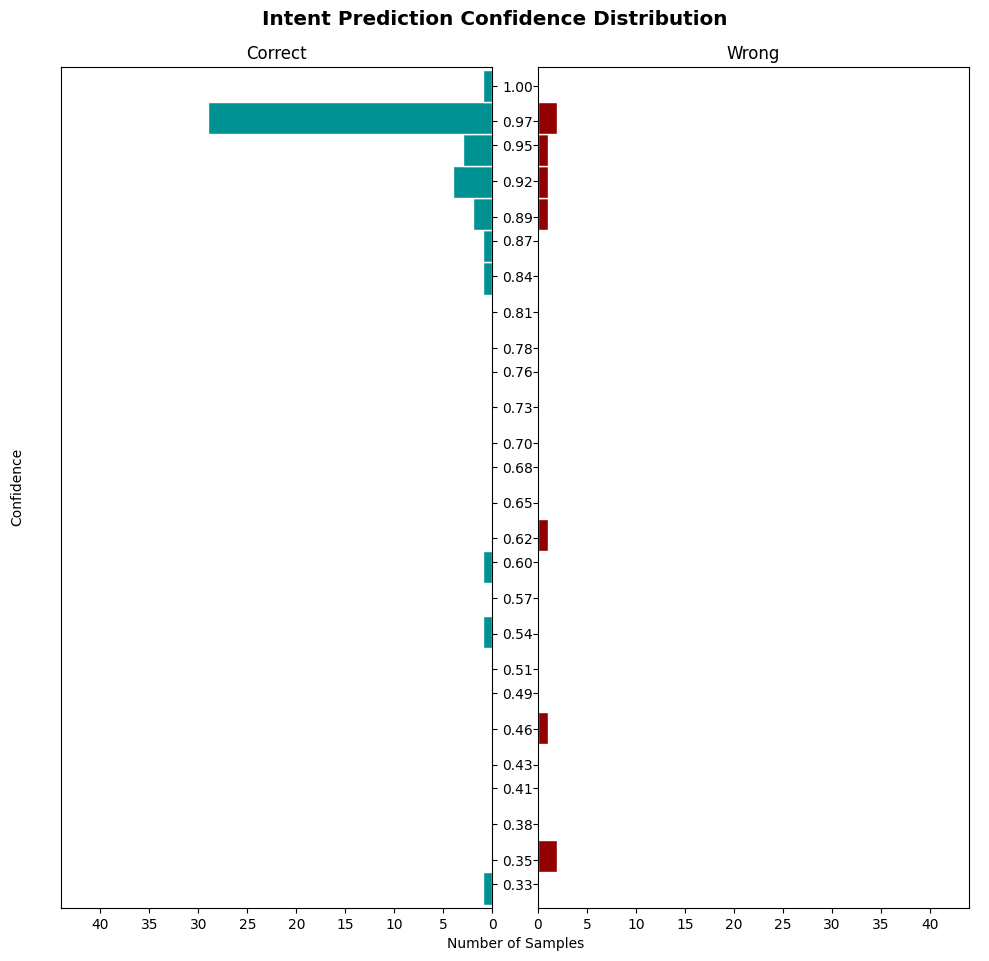}
\caption{Pipeline P8 - Intent Histogram}
\label{p8intent}
\end{figure}

\noindent In aim for improving test score performance, we added 2 components with pipeline \textbf{P1} to build pipeline \textbf{P3}. They are  \textbf{(1)} EntitySynonymMapper: maps synonymous entity values to the same values, and \textbf{(2)} FallbackClassifier: used to give a default response for user questions with intents out of scope. In Table \ref{results}, we can see that test score performance gets better than pipeline \textbf{P1}. \par

\noindent We add a custom fastTextFeaturizer specifically built for Bangla Language with pipeline\textbf{P3} for pipeline \textbf{P4} to get better results. fastText featurizer returns dense word vectors trained using CBOW\cite{mikolov2013efficient} with position-weights on a huge Bangla corpus. As we see in Table \ref{results}, the scores in all 4 metrics decrease. \par

\noindent To investigate the reason behind the decline in performance in pipeline \textbf{P5}, we take out the regex featurizer and run the experiment again and see a sharp rise in all the metrics. Here we come to a conclusion that regex featurizer does not work well with fastText word vectors; this is because regex sparse vectors nullify fastext character word models when concatenated together which results in better performance. \par

\noindent If pipeline \textbf{P6} is compared with pipeline \textbf{P5}, we can see the only difference is in the tokenizer component. \textbf{P6} uses our custom Bangla tokenizer but \textbf{P5} does not, and consequently \textbf{P6} produces better results for all the metrics. \par

\noindent For pipeline \textbf{P7}, we replace our custom Bangla tokenizer with BERT \cite{devlin2018bert} language model tokenizer and we also replace fastTextFeaturizer with BERT language model featurizer. Both of these components are pretrained on a huge Bangla corpus. And we can see slightly better scores than pipeline \textbf{P6}. \par

\noindent Finally, in pipeline \textbf{P8}, we add RegexFeaturizer with pipeline \textbf{P7}. We can see in Table \ref{results} that we get the best result in this setup. The DIET architecture concatenates sparse features from regex featurizer and dense features from BERT language model featurizer. Featurizers in the same pipeline need to generate similar types of features so that they do not nullify each other. From this intuition, we can derive that regex featurizer works better with BERT language model featurizer because both of them deals with defined subwords or patterns. 

\subsection{Result Analysis}
\noindent According to Table \ref{pipelines} and Table \ref{results}, we can conclude that Pipeline P6 works best with fastText dense featurizer and Pipeline P8 works best with BERT language model featurizer. In this section, we compare the results from Pipeline P6 and P8 so that we can describe the difference more visually. \par 

\noindent If we compare the confusion matrix of Pipeline P8 in Figure \ref{p8confmat} with Pipeline P6 in Figure \ref{p8confmat}, we can clearly see that the sum of diagonal in Pipeline P6 Confusion Matrix is lower than the sum of diagonal Pipeline P8. Pipeline P6 is getting confused between course\_access\_duration and course\_details while Pipeline P8 can successfully identify the difference between them. If we look more closely, we can find more anomalies like this. \par

\noindent We can also compare the intent histograms in Figure \ref{p6intent} and \ref{p8intent}. Here, we can find that Pipeline P8 is more confident when they are providing correct responses which is not the case in Pipeline P6. In Pipeline P6, the confidence varies a lot when the response is correct. On the other hand, Pipeline P6 is more confident when they are providing wrong responses. In the case of pipeline P8, we can notice inconsistent confidence values for wrong responses. \par

\noindent Finally, comparing pipeline P6 and P8 we found that P8 performs better because it uses Language-agnostic BERT Sentence Embedding (LaBSE) \cite{feng2020language} which is a powerful tool to encode texts from different languages into a shared embedding space with high efficiency. LaBSE is trained using large multilingual corpus like Common Crawl and Wikipedia. This setup can also provide competitive performance in other languages since it is based on a language-agnostic state-of-the-art sentence embedding. 

\section{Conclusion and Future works}

\noindent The default components of Rasa Framework perform poorly for Bangla as it is still a low-resource language. Hence the need for custom components built specifically for Bangla arises. In our experiments, we showed a detailed comparative analysis of the effects of each individual default and custom component.  The dataset we collected, annotated, and reviewed was imbalanced. So, we also needed to deal with the imbalanced class problem. \par

\noindent In the future, we plan to extend and improve the quality of our existing dataset by collecting more data and going through more rigorous reviewing. We also plan to add more custom components to the Rasa pipeline, for example, state-of-the-art transformer models, state-of-the-art multilingual featurizers, and many others. Furthermore, we also have plans to build a multilingual chatbot that can interact with users of different languages from different countries and cultures around the globe.

\bibliographystyle{ieeetr}
\bibliography{main.bib}

\end{document}